\documentclass[preprint,12pt]{article}

\usepackage{style}
\usepackage{times}
\usepackage{url}
\usepackage{latexsym}

\usepackage{acro}
\usepackage{multirow}
\usepackage{array}
\usepackage{xcolor}
\usepackage{authblk}
\usepackage{graphicx}

\newcommand{\new}[1]{\textcolor{blue}{#1}}

\title{Comparative Evaluation of \\ Machine Translation Systems on
Images with Text}

\date{}

\DeclareAcronym{art}{short = ART, long  = approximate randomization testing, cite = Riezler05}
\DeclareAcronym{beer}{short = BEER, long = BEtter Evaluation as Ranking, cite = stanojevic2014beer}
\DeclareAcronym{bleu}{short = BLEU, long  = bilingual evaluation understudy, cite = papineni2002bleu}
\DeclareAcronym{bwer}{short = bWER, long = bag-of-words WER, cite = Vidal2023bwer}
\DeclareAcronym{bpe}{short = BPE, long  = byte pair encoding, cite = Gage94}
\DeclareAcronym{cbmt}{short = CBMT, long  = character-based machine translation}
\DeclareAcronym{cbsmt}{short = CBSMT, long  = character-based statistical machine translation}
\DeclareAcronym{cbnmt}{short = CBNMT, long  = character-based neural machine translation}
\DeclareAcronym{chrf}{short = chrF, long = character n-gram F-score, cite = popovic2015chrf}
\DeclareAcronym{cm}{short = CM, long  = confidence measures}
\DeclareAcronym{cnn}{short = CNN, long = convolutional neural network}
\DeclareAcronym{craft}{short = CRAFT, long = Character Region Awareness for Text Detection, cite = baek2019craft}
\DeclareAcronym{crnn}{short = CRNN, long = convolutional recurrent neural network}
\DeclareAcronym{cer}{short = CER, long  = Character Error Rate}
\DeclareAcronym{cv}{short = CV, long = computer vision}
\DeclareAcronym{dr}{short = DR, long = document recognition}
\DeclareAcronym{fast}{short = FAST, long = Faster Arbitrarily-Shaped Text Detector, cite = chen2023fast}
\DeclareAcronym{fda}{short = FDA, long  = feature decay algorithm, cite = Biccici15}
\DeclareAcronym{hplt}{short = HPLT, long = High Performance Language Technologies, cite = gilbert2024new}
\DeclareAcronym{htr}{short = HTR, long  = handwritten text recognition}
\DeclareAcronym{il}{short = IL, long  = incremental learning}
\DeclareAcronym{imt}{short = IMT, long  = image-based machine translation}
\DeclareAcronym{iimt}{short = IIMT, long = in-image machine translation}
\DeclareAcronym{inmt}{short = INMT, long  = interactive neural machine translation}
\DeclareAcronym{ismt}{short = ISMT, long  = interactive statistical machine translation}
\DeclareAcronym{ksr}{short = KSR, long  = key stroke rate}
\DeclareAcronym{llm}{short = LLM, long  = large language model}
\DeclareAcronym{mllm}{short = MLLM, long  = multimodal large language model}
\DeclareAcronym{MoE}{short = MoE, long = Mixture-of-Experts }
\DeclareAcronym{lstm}{short = LSTM, long  = long short-term memory, cite = Hochreiter97}
\DeclareAcronym{mar}{short = MAR, long  = mouse action rate}
\DeclareAcronym{mt}{short = MT, long  = machine translation}
\DeclareAcronym{mmt}{short = MMT, long = multimodal machine translation}
\DeclareAcronym{nlp}{short = NLP, long  = natural language processing}
\DeclareAcronym{nmt}{short = NMT, long  = neural machine translation}
\DeclareAcronym{ocr}{short = OCR, long = optical character recognition}
\DeclareAcronym{ood}{short = OOD, long = out-of-distribution, cite = hendrycks2016ood}
\DeclareAcronym{pe}{short = PE, long  = post-editing}
\DeclareAcronym{rbmt}{short = RBMT, long  = rule-based machine translation}
\DeclareAcronym{relu}{short = ReLU, long  = rectified linear unit}
\DeclareAcronym{rlhf}{short = RLHF, long = reinforcement learning with human feedback, cite = lambert2025rlhf}
\DeclareAcronym{rnn}{short = RNN, long  = recurrent neural network, cite = {Hochreiter97}}
\DeclareAcronym{sgd}{short = SGD, long  = stochastic gradient descend}
\DeclareAcronym{smt}{short = SMT, long  = statistical machine translation}
\DeclareAcronym{ter}{short = TER, long  = translation error rate, cite = Snover2006ter}
\DeclareAcronym{timt}{short = TIMT, long = text-image machine translation, cite = lan2023exploring}
\DeclareAcronym{wer}{short = WER, long  = Word Error Rate}
\DeclareAcronym{wsr}{short = WSR, long  = word stroke rate}
\DeclareAcronym{xml}{short = XML, long  = eXtensible Markup Language}

\begin{document}


\author[1]{Blai Puchol}
\author[1]{Sergio G\'omez Gonz\'alez}
\author[1,2]{Miguel Domingo}
\author[1,2]{Francisco Casacuberta}

\email{bpucsal@inf.upv.es \\ \{sgomgon,midobal,fcn\}@prhlt.upv.es}

\affil[1]{PRHLT Research Center - Universitat Polit{\`e}cnica de Val{\`e}ncia \\
Camino de Vera, s/n - Valencia}
\affil[2]{ValgrAI - Valencian Graduate School and Research Network for Artificial Intelligence\\Camino de Vera s/n}

\maketitle
\begin{abstract}
    This work presents a comparative evaluation of machine translation systems applied to images containing textual information, a task that lies at the intersection of computer vision and natural language processing. The study compares three main paradigms: modular pipelines that separate text detection, recognition, and translation; multi-modal large language models (MLLMs) capable of processing both image and text jointly; and an end-to-end model, Translatotron-V, which directly generates translated images. The modular systems employ state-of-the-art OCR (docTR) combined with multilingual LLMs such as Llama and EuroLLM, while the evaluated MLLMs include different configurations of Gemini 2.5. Experiments were conducted on parallel multilingual datasets covering multiple language pairs, with evaluation based on BLEU, chrF, and TER metrics. The results show that modular pipelines outperform the end-to-end approach, while MLLMs achieve the best overall performance, demonstrating superior flexibility and contextual understanding. These findings underscore the effectiveness of multi-modal reasoning for image-to-text translation and provide a solid foundation for future research on integrating visual understanding and language generation in multilingual settings.
\end{abstract}

\section{Introduction}
\Ac{mt} was born as an automated way to translate text from one language to another, allowing more readers to understand the information contained in the original text. 
However, most of the information that humans produce and receive is visual; yet it contains text in some language. As a result, people who do not speak the language cannot understand most of the content in these images. In order to ensure the accessibility of the visually codified text, \ac{timt} has emerged. This research field includes solutions dedicated to obtaining textual translations of images containing embedded text.

Since \ac{timt} models receive an image and produce text, the task is at the intersection between \ac{cv} and \ac{nlp}. Specifically, \ac{timt} can be expressed as a combination of \ac{ocr} and \ac{mt}. From an \ac{ocr} point of view, \ac{timt} solutions should locate and extract the text embedded in the images. However, from an \ac{mt} perspective, they must adequately translate the recognized text into the target language. Sometimes, this division is more evident when different models are used in a cascade-model \cite{qian2024anytrans} or end-to-end \cite{ma2022improving} fashion. 

In this work, we propose a modular pipeline that allows the use of different models to perform sequentially \ac{ocr} and \ac{mt}. Furthermore, we have selected a dataset \cite{lan2024translatotron} that was previously used in other \ac{timt} works. Thus, we tested our pipeline with different models to find the best combination. Moreover, we compared it with an end-to-end model \cite{lan2024translatotron} and some \ac{mllm} to assure its potential. We found that our pipeline is competitive with selected \ac{mllm}s, and surpasses the performance of the end-to-end model.

The rest of this paper is structured as follows: first, we present an overview of the state of the art with current tendencies in the \ac{timt} field, and the best performing models. After that, we describe the methodology that we have followed in our experiments in Section \ref{sec:methodology}. There, we introduce our modular pipeline (Section \ref{sec:pipeline}), an end-to-end system, and some \ac{mllm}s (Section \ref{sec:models}). Then, in Section \ref{sec:experiments}, we present the evaluation framework that we have followed. Afterward, in Section \ref{sec:results}, we review and study the results of our experimental session. Finally, the article ends with our conclusions and future steps in Section \ref{sec:conclusion}.

\section{Related Work}
\Ac{imt} is considered a sub-field of \ac{mmt} dedicated to the translation of textual content embedded in the media \cite{elliott2017imagination,gao2025multimodal}. Additionally, \ac{imt} can be divided into two sub-tasks: \ac{timt} and \ac{iimt}. The objective of the first is to translate the textual content of an image, obtaining a pure textual output \cite{lan2025towards,ma2023modal}. Meanwhile, \ac{iimt} aims to substitute the text in the image for its translation into another language \cite{tian2025exploring,tian2025prim}. Thus, the result is a new image with the same visual components but a rendered translation of the text instead of the original.

To address the task of \ac{timt}, multiple proposals have been made, ranging from cascade-based modular pipelines \cite{qian2024anytrans} to end-to-end models \cite{tian2023image,lan2023exploring,liang2024document,lan2024translatotron}. The former often divides the task into \ac{ocr}, followed by \ac{mt} on the recognized text. They tend to suffer from error propagation \cite{tian2023image}, but can exploit highly optimized models trained on vast amounts of data. In contrast, end-to-end models face the task as a whole solid challenge to avoid error propagation. They are highly specialized models that are trained only for \ac{timt} with the available data.

The common approach to cascade pipelines starts with detecting and recognizing text in images. While vanilla \ac{ocr} models focus primarily on scanned documents, \cite{li2023trocr,cheng2017focusing} \ac{timt} images usually do not follow this pattern. Real world applications of \ac{timt} require detecting text in natural scenes. This is a more challenging task, since images usually contain complex backgrounds, variable lighting, distortions, arbitrary orientations, and diverse fonts. To face this type of challenge, the availability of data is vital; therefore, datasets were created incorporating these special conditions \cite{gupta2016synthetic,veit2016coco,wang2011end}. Furthermore, challenges such as the one organized at \emph{ICDAR 2015} \cite{karatzas2015icdar} promote performance improvement in natural-scene \ac{ocr} models.

After recognizing text in images, cascade approaches tend to use an \ac{mt} model to obtain the final output \cite{qian2024anytrans}. \Ac{mt} has a long history with a large availability of models, both direction-specific and multilingual \cite{hameed2025survey}. In addition, conferences such as \emph{WMT} \cite{chatterjee2019findings,farajian2020findings,wang2024findings} promote research on several open challenges of \ac{mt}, such as quality estimation \cite{specia2021findings}. Many datasets have been created to address various specific \ac{mt} tasks \cite{koehn2005europarl,tiedemann2020opus}, with increasing difficulty. Lately, the use of \ac{llm} has also been introduced in \ac{mt}, even with the support of institutions such as the European Union \cite{martins2025eurollm}.

The other common approach to \ac{timt} is to train an end-to-end model. This branch is closely related to \ac{iimt} since \ac{timt} is usually used as an auxiliary task to obtain an \ac{iimt} model \cite{lan2024translatotron}. Attempting to directly attack the problem at the pixel level \cite{tian2023image} has proven to be a poor approach \cite{lan2023exploring}. Therefore, other approaches have tried to involve a codebook in their models architecture \cite{lan2023exploring,lan2024translatotron}. Other models produce a formatted output to render the image in case of need \cite{liang2024document}.

Usually, given the scarcity of large-scale high-quality parallel datasets \cite{li2025mit10m}, end-to-end models are trained with synthetic data \cite{lan2024translatotron}. To our knowledge, the only manually curated datasets that exist to date are \emph{OCRMT30k} \cite{lan2023exploring}, \emph{DoTA} \cite{liang2024document} and \emph{ECOIT} \cite{zhu2023peit}. These datasets are valuable; however, they are small in scale and restricted in language coverage. 

The revolution of \ac{llm}s has also affected \ac{timt} tasks, since they are often used as part of pipelines. Tools such as \emph{Anytrans} \cite{qian2024anytrans} include these models as translators. They can help mitigate \ac{ocr} errors through contextual understanding \cite{wang2025rethinking}.

\section{Methodology}\label{sec:methodology}

This section describes the experimental framework used to evaluate \ac{imt} systems. We compare modular pipelines that separate \ac{ocr} from text translation with end-to-end systems. All models are evaluated on a multilingual dataset of synthetic images that contain rendered text, ensuring consistent conditions across architectures. The methodology covers the dataset, the modular pipeline, the evaluated models, and the prompting strategies used for each \ac{llm} and \ac{mllm}.

\subsection{Corpora}

\begin{figure}[ht]
    \centering
    \begin{minipage}[b]{0.18\textwidth}
        \centering
        \includegraphics[width=\textwidth]{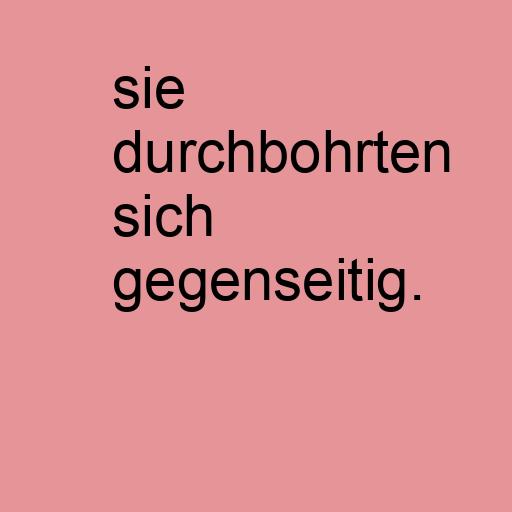}
    \end{minipage}
    \begin{minipage}[b]{0.18\textwidth}
        \centering
        \includegraphics[width=\textwidth]{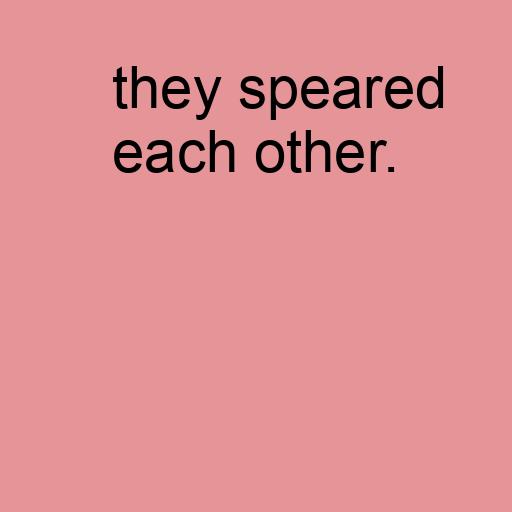}
    \end{minipage}

    \caption{Example of an image pair. }
    \label{fig:pair_example}
\end{figure}

All experiments use the multilingual dataset introduced by \citet{lan2024translatotron}, designed specifically for \ac{iimt}. The dataset consists of synthetic images generated by rendering parallel text from IWSLT14 (German{\textendash}English) and IWSLT17 (French{\textendash}English, Romanian{\textendash}English); hereafter, De{\textendash}En, Fr{\textendash}En and Ro{\textendash}En, respectively. Each pair of sentences is placed on $512\times512$ images using the Python Pillow library, with black Arial font and random variations in position, rotation, and background color. An example can be seen in Figure \ref{fig:pair_example}. This setup introduces moderate visual noise while maintaining readability. 

Table \ref{tab:dataset_stats} summarizes the number of samples per language pair and partition. All images contain aligned text in two languages{\textemdash}enabling direct comparison between modular and end-to-end approaches under identical conditions.

\begin{table}[ht]
    \centering
    \begin{tabular}{cccc}
    \hline
    \textbf{Language pair} & \textbf{Train} & \textbf{Validation} & \textbf{Test} \\
    \hline
    \textbf{De--En} & $81\,741$ & $3\,765$ & $3\,527$ \\
    \textbf{Fr--En} & $131\,707$ & $448$ & $4\,927$ \\ 
    \textbf{Ro--En} & $123\,671$ & $445$ & $1\,008$\\
    \hline
    \end{tabular}
    \caption{Number of pairs of images per language pair and split in the dataset.}
    \label{tab:dataset_stats}
\end{table}

\subsection{Modular pipeline}\label{sec:pipeline}

The modular pipeline processes input images by first extracting text through \ac{ocr} and then translating it into the target language.

\subsubsection{OCR Stage}

Two open-source \ac{ocr} systems were considered for the text extraction stage: \emph{EasyOCR}\footnote{\url{https://github.com/JaidedAI/EasyOCR}} and \emph{docTR}\footnote{\url{https://github.com/mindee/doctr}} \cite{doctr2021}. 

\emph{EasyOCR} employs a deep learning pipeline combining a \ac{craft} model for detection with a \ac{crnn} \cite{shi2016end} for recognition. Although it supports more than 80 languages and offers high inference speed, its robustness can be degraded under significant text distortion.

\emph{docTR} provides a highly flexible framework built on TensorFlow and PyTorch, offering a choice of different detection and recognition models. It is particularly known for its high accuracy and strong performance in handling rotated or multi-line text.

After a preliminary accuracy comparison \new{(Section~\ref{sec:ocr-eval})}, \emph{docTR} was selected for all the main experiments due to its superior performance. Leveraging its modularity, we set up a specific two-stage pipeline: for the text detection stage, we employed the \ac{fast}, and for the text recognition stage, we chose a \ac{cnn}. This recognition model uses a VGG16-BN backbone, consisting of a VGG16 \cite{simonyan2014very} that adds batch normalization layers, which act as a powerful visual feature extractor for cropped text images. These features are then sequentially fed into the \ac{rnn}, which decodes them into the final character sequence.

\subsubsection{Translation Stage}

At this stage, the text extracted in the previous phase is fed into the selected \ac{mt} models. Our methodology distinguishes between two fundamentally different approaches to this task, based on the model's architecture.

For standard multilingual \ac{nmt} systems, such as \textit{M2M100-1.2B} \cite{fan2021beyond}, the process is direct. These models are specialized exclusively for translation, and the task is conditioned by providing predefined source and target language tags to the input \cite{fan2021beyond}.

In contrast, \ac{llm}s are general-purpose instruction-following models. They are not specialized for a single task and must be guided to perform the translation. This is achieved by formatting the input via natural-language instructions, i.e., a \textit{prompt}. These prompts, further detailed in Section~\ref{prompting}, explicitly request only the translated text to ensure a clean, comparable output.

This modular setup allows separate evaluation of \ac{ocr} and translation quality using dedicated metrics.

\subsection{Models}\label{sec:models}

We considered four main categories of models, selected to provide a clear contrast in architecture, specialization and capability. 

As a traditional NMT  \cite{mohammadshahi2022-mall,boyd2025machine}, we selected \textbf{M2M100-1.2B} \cite{fan2021beyond}. This model employs a standard Transformer encoder-decoder architecture and is known for its strong cross-lingual translation performance.

For the \textbf{LLM-based modular pipeline}, we employed general-purpose, decoder-only models. This included instruction-tuned models from Meta’s Llama family\footnote{\url{https://www.llama.com/}}, version 3 \cite{dubey2024llama}: \textbf{Llama-3.2-1B-Instruct}, \textbf{Llama-3.2-3B-Instruct}, and \textbf{Llama-3.1-8B-Instruct}.
We also included \textbf{EuroLLM-1.7B} and \textbf{EuroLLM-9B} \cite{martins2025eurollm}, trained with a focus on European languages. This selection allows us to compare the general pre-training centered on English with the linguistic specialization of EuroLLM. Moreover, we can assess the model's scaling impact.

For the \textbf{multi-modal} category, we tested three configurations of \textbf{Gemini-2.5} \cite{comanici2025gemini}: \textit{flash-lite}, \textit{flash}, and \textit{pro}. These models are native multi-modal and utilize a sparse \ac{MoE} architecture for efficient scaling. Thus, this architecture functions as an end-to-end system, processing the raw image and text prompt in a single step, thus avoiding the separate \ac{ocr} stage entirely.

Finally, we compared our results with an \textbf{end-to-end image-to-image model}, \textbf{Translatotron-V} \cite{lan2024translatotron}, using accuracy values obtained through private communication with the authors. This provides a second, more specialized comparison for end-to-end \ac{iimt} performance, although the results were limited to two language pairs due to the inaccessibility of a public checkpoint.

\subsection{Prompting Strategies}
\label{prompting}

Prompt design was a critical step in obtaining consistent and clean translations of \ac{llm}s and \ac{mllm}s. As demonstrated in previous work \cite{liu2023pre}, well-structured prompts significantly improve model behavior in translation tasks by reducing irrelevant or reformulated outputs. Our primary goal was to engineer prompts that enforced a minimal response, thereby minimizing the need for post-processing and ensuring a fair, metric-based comparison across model families.

The required prompt structure varied depending on each model's architecture. For EuroLLM, a simple format that specifies the source language and target language tags was sufficient to obtain clean translations. In contrast, LLaMA models required more explicit instructions to suppress unnecessary commentary. We experimented with both direct text commands (e.g., ``Respond only with the translation'') and structured JSON outputs to facilitate automated parsing. For multi-modal Gemini models, prompts were designed to integrate both \ac{ocr} and translation into a single query, instructing the model to extract the source text from the image and then to return only the final translation.

Table~\ref{tab:prompts_models} provides concrete examples of the prompts used for each model. This careful control was essential for standardizing the outputs for evaluation.

\begin{table}[ht]
\centering
\begin{tabular}{m{0.21\linewidth}|m{0.67\linewidth}}
\hline
\textbf{Model} & \textbf{Prompt template} \\
\hline
\textbf{EuroLLM} &
\texttt{English: Some text} \newline
\texttt{German:} \\
\hline
\textbf{Llama \newline (to text)} &
\texttt{Translate the following into German. Respond only with the translation, no comments:} \newline
\texttt{English: Some text} \newline
\texttt{German:} \\
\hline
\textbf{Llama \newline (to JSON)} &
\texttt{Translate into French. Respond only in this JSON format: {"translation": "<text>"}} \newline
\texttt{English: "Some text"} \newline
\texttt{French (JSON):} \\
\hline
\textbf{Gemini} &
\texttt{Extract the English text from this image and translate it to French. Provide only the final French translation.} \\
\hline
\end{tabular}
\caption{Examples of prompts used for different models in our experiments.}
\label{tab:prompts_models}
\end{table}

\section{Experimental Session}
\label{sec:experiments}

This section describes the technical framework used to perform our evaluation. We first detail the implementation, including the software libraries and hardware used to run the different model families. We then define the specific evaluation metrics used to assess the performance of both \ac{ocr} and translation stages.

\subsection{Implementation Details}

Our experiments were conducted using two primary methods, depending on model accessibility. All open-weight models (M2M100, Llama, and EuroLLM) were run locally using the Hugging Face \texttt{transformers} library \cite{wolf2020transformers}. These tests were performed on servers equipped with \textit{NVIDIA RTX 4090} and \textit{NVIDIA Quadro RTX 8000} GPUs.

For the closed-source Gemini family, the models were evaluated using the official Google GenAI API, powered by Google Cloud.

\subsection{Evaluation Metrics}
We assessed both the OCR and translation stages using complementary metrics.  
For OCR, we report \ac{cer} and \ac{wer}~\cite{morris2004and}, which quantify transcription errors at the character and word levels, respectively. Lower values indicate better recognition accuracy.

For translation quality, we used three standard metrics widely adopted in \ac{mt} research:
\begin{itemize}
    \item \Ac{bleu}: Measures $n$-gram overlap between model output and reference translations.
    \item \Ac{chrf}: Computes an F-score based on character $n$-gram precision and recall, capturing fine-grained similarity.
    \item \Ac{ter}: Calculates the normalized number of edits required to match the reference translation.
\end{itemize}

For consistency, \ac{mt} metrics  were computed using \texttt{sacreBLEU} \cite{post2018call}. Additionally, we applied \ac{art}{\textemdash}with $10,000$ repetitions and using a $p$-value of $0.05${\textemdash}to determine whether two systems presented statistically significance.

\section{Results}\label{sec:results}

This section presents the empirical results of the experiments detailed in Section~\ref{sec:experiments}. We start by reporting the outcome of our preliminary \ac{ocr} evaluation, which informed the configuration of our modular pipeline. Then, we present and analyze the main translation quality results, comparing the performance between architectures.

\subsection{OCR Evaluation}\label{sec:ocr-eval}

To identify the best OCR module for our pipeline, we compared \texttt{EasyOCR} and \texttt{docTR} on the test partition using WER and CER metrics (Table~\ref{tab:ocr-comparison}). The results show that \texttt{docTR} consistently outperformed \texttt{EasyOCR} in all languages, reducing the error rates by up to $13.9$ percentage points in WER and $12.5$ in CER. These improvements confirmed \texttt{docTR} as the preferred OCR component for subsequent translation experiments.

\begin{table}[ht]
\centering
\begin{tabular}{c cc cc}
\hline
\multirow{2}{*}{}  &
\multicolumn{2}{c}{\textbf{WER $\downarrow$}} & 
\multicolumn{2}{c}{\textbf{CER $\downarrow$}} \\ 
\cline{2-5}
& \textbf{EasyOCR} & \textbf{docTR} & \textbf{EasyOCR} & \textbf{docTR} \\ 
\hline
\textit{De} & $17.3$ & \textbf{$11.0$} & $12.5$ & \textbf{$3.0$} \\
\textit{En} & $21.8$ & \textbf{$7.9$} & $15.6$ & \textbf{$3.1$} \\
\hline
\textit{Fr} & $30.6$ & \textbf{$19.0$} & $16.4$ & \textbf{$4.7$} \\
\textit{En} & $32.7$ & \textbf{$20.2$} & $18.3$ & \textbf{$6.3$} \\
\hline
\end{tabular}
\caption{Comparison of OCR results between \texttt{EasyOCR} and \texttt{docTR}.}
\label{tab:ocr-comparison}
\end{table}

\subsection{Translation results}
\label{translation-results}

In this section, we analyze the translation performance of all tested models across the four language pairs: English--German, German--English, English--French, and French--English. Recall that all models were evaluated on the same test partition.

Across all language pairs, the results reveal several consistent patterns. Traditional \ac{nmt} models provide solid baselines, but are generally outperformed by larger \ac{llm}s and MLLMs. Specifically, as seen in Table~\ref{table:en-de-scores}, the Llama models show a clear improvement with scale: smaller models perform notably below the \ac{nmt} system, but the 8B version approaches or improves its performance. The EuroLLM family further improves translation quality, with EuroLLM-9B exceeding the \ac{nmt} scores in all metrics evaluated for English--German. Multi-modal Gemini models outperform all other systems, and the most advanced configuration, Gemini-2.5-pro, achieves the best BLEU, chrF, and TER values overall. The end-to-end model Translatotron-V, available only for this language pair, performs significantly worse, with BLEU values of around 17\%, suggesting that direct image-to-image translation remains a challenging task.

\begin{table}[ht]
\centering
\begin{tabular}{l|ccc}
\hline
\textbf{Model} & \textbf{BLEU $\uparrow$} & \textbf{chrF $\uparrow$} & \textbf{TER $\downarrow$} \\
\hline
M2M100-1.2B & $27.7\dagger$ & $53.3\dagger$ & $62.3\dagger$ \\
Llama-3.2-1B & $18.2$ & $44.3$ & $75.2$ \\
Llama-3.2-3B & $23.7$ & $50.5$ & $67.3$ \\
Llama-3.1-8B & $26.8\dagger$ & $53.2\dagger$ & $63.0\dagger$ \\
EuroLLM-1.7B & $26.9\dagger$ & $53.1\dagger$ & $63.5\dagger$ \\
EuroLLM-9B & $31.2$ & $56.4$ & $58.4$ \\
Gemini-2.5-lite & $32.1$ & $57.3$ & $58.0$ \\
Gemini-2.5-pro & \textbf{$35.6$} & \textbf{$59.9$} & \textbf{$52.5$} \\
Translatotron-V* & $17.9$ & – & – \\
\hline
\end{tabular}
\caption{English–German translation results. All results are statistically different except those denoted by $\dagger$. *Translatotron-V values were obtained via private communication with the authors.}
\label{table:en-de-scores}
\end{table}

For the opposite direction (German--English), shown in Table~\ref{table:de-en-scores}, a similar trend can be observed. Although the \ac{nmt} baseline yields relatively low results, LLM-based systems demonstrate substantial improvements, particularly at larger scales. EuroLLM-9B achieves a BLEU score of approximately 39 points and a TER close to 48 points, marking a strong gain in both \ac{nmt} and smaller \ac{llm}s. Once again, Gemini models dominate this task, with the Gemini-2.5-pro configuration yielding the highest performance across all metrics. No reference results were available for Translatotron-V in this direction, limiting the comparison to text-based and multi-modal approaches.

\begin{table}[ht]
\centering
\begin{tabular}{l|ccc}
\hline
\textbf{Model} & \textbf{BLEU $\uparrow$} & \textbf{chrF $\uparrow$} & \textbf{TER $\downarrow$} \\
\hline
M2M100-1.2B & $19.1$ & $40.5$ & $70.9$ \\
Llama-3.2-1B & $25.6$ & $47.8$ & $62.5$ \\
Llama-3.2-3B & $30.1$ & $52.5$ & $57.8$ \\
Llama-3.1-8B & $34.2\dagger$ & $55.9\dagger$ & $53.4\dagger$ \\
EuroLLM-1.7B & $33.9\dagger$ & $54.9\dagger$ & $53.7\dagger$ \\
EuroLLM-9B & $39.1$ & $59.5$ & $48.1$ \\
Gemini-2.5-lite & $35.5$ & $56.7$ & $51.5$ \\
Gemini-2.5-flash & $38.3$ & $58.0$ & $52.6$ \\
Gemini-2.5-pro & \textbf{$41.4$} & \textbf{$61.2$} & \textbf{$47.2$} \\
\hline
\end{tabular}
\caption{German--English translation results. All results are statistically different except those denoted by $\dagger$.}
\label{table:de-en-scores}
\end{table}

In the English-French direction (Table~\ref{table:en-fr-scores}), the \ac{nmt} baseline achieves competitive results, but smaller \ac{llm}s lag behind it. As before, scaling up model size leads to marked gains: both EuroLLM-9B and the larger Llama variant surpass the baseline, indicating the benefits of capacity and multilingual specialization. Among \ac{mllm}s, all Gemini configurations outperform text-only systems, with the Gemini-2.5-pro model achieving, again, the highest scores in BLEU, chrF, and TER. Translatotron-V was not evaluated for this pair, so the results focus on modular and multi-modal systems.

\begin{table}[ht]
\centering
\begin{tabular}{l|ccc}
\hline
\textbf{Model} & \textbf{BLEU $\uparrow$} & \textbf{chrF $\uparrow$} & \textbf{TER $\downarrow$} \\
\hline
M2M100-1.2B & $34.7$ & $57.7$ & $55.0$ \\
Llama-3.2-1B & $22.4$ & $46.5$ & $67.8$ \\
Llama-3.2-3B & $27.5$ & $52.5$ & $60.1$ \\
Llama-3.1-8B & $29.3$ & $55.5$ & $55.8$ \\
EuroLLM-1.7B & $33.6$ & $56.0$ & $55.7$ \\
EuroLLM-9B & $36.8$ & $59.9$ & $50.1$ \\
Gemini-2.5-lite & $40.5$ & $62.2$ & $47.3$ \\
Gemini-2.5-flash & $37.3$ & $57.2$ & $55.1$ \\
Gemini-2.5-pro & \textbf{$44.6$} & \textbf{$65.2$} & \textbf{$44.1$} \\
\hline
\end{tabular}
\caption{English--French translation results. All results are statistically different.}
\label{table:en-fr-scores}
\end{table}

Finally, for the French--English translation task (Table~\ref{table:fr-en-scores}), the \ac{nmt} model shows its weakest performance among the studied language pairs. Consequently, all \ac{llm}s outperform it, even at smaller scales. EuroLLM-9B leads among text-based systems, while the Gemini series once again dominates, with Gemini-2.5-pro reaching the best overall results across all metrics. Interestingly, in this direction, the Translatotron-V model performs better than the \ac{nmt} baseline, achieving a BLEU score of around 34, although it still falls short compared to \ac{llm}s and \ac{mllm}s.

\begin{table}[ht]
\centering
\begin{tabular}{l|ccc}
\hline
\textbf{Model} & \textbf{BLEU $\uparrow$} & \textbf{chrF $\uparrow$} & \textbf{TER $\downarrow$} \\
\hline
M2M100-1.2B & $25.2$ & $46.1$ & $65.3$ \\
Llama-3.2-1B & $32.3\ddagger$ & $53.7\ddagger$ & $54.8\ddagger$ \\
Llama-3.2-3B & $32.9\ddagger$ & $55.0\ddagger$ & $52.9\ddagger$ \\
Llama-3.1-8B & $35.3\dagger$ & $57.1\dagger$ & $48.0\dagger$ \\
EuroLLM-1.7B & $35.7\dagger$ & $57.2\dagger$ & $49.1\dagger$ \\
EuroLLM-9B & $37.6$ & $59.0$ & $45.3$ \\
Gemini-2.5-lite & $39.5$ & $59.2$ & $46.6$ \\
Gemini-2.5-flash & $42.2$ & $61.1$ & $45.4$ \\
Gemini-2.5-pro & \textbf{$43.5$} & \textbf{$62.6$} & \textbf{$43.1$} \\
Translatotron-V* & $34.2$ & – & – \\
\hline
\end{tabular}
\caption{French--English translation results. All results are statistically different except those denoted by $\dagger$ and $\ddagger$ respectively. *Values obtained via private communication.}
\label{table:fr-en-scores}
\end{table}

In summary, comparative analysis across all four translation directions reveals a consistent performance hierarchy: traditional \ac{nmt} systems establish reliable baselines; large-scale LLMs, especially EuroLLM, provide significant improvements; and multi-modal models, particularly Gemini-2.5-pro, deliver the highest translation quality overall. These findings underline the growing advantage of multi-modal architectures that jointly process visual and linguistic information. Nevertheless, the lack of complete results for Translatotron-V highlights ongoing challenges and the need for more accessible end-to-end image-to-image translation benchmarks.

\section{Limitations}

This study has several limitations that should be considered when interpreting the results. First, the evaluation is based on a synthetic dataset where text is rendered under controlled conditions. Although this setup enables clean comparisons across systems, it can introduce biases and may overestimate performance relative to real-world scenarios. In practice, text in natural images is often affected by motion blur, occlusion, perspective distortion, complex backgrounds, and uncontrolled lighting. Therefore, evaluating with in-the-wild images would provide a more realistic estimate of system robustness.

Second, language coverage remains limited to a small set of relatively high-resource European language pairs. The conclusions may not transfer directly to typologically distant languages, non-Latin scripts, or genuinely low-resource settings where OCR and MT errors can compound differently.

Third, the analysis relies primarily on automatic metrics (BLEU, chrF, and TER). Although standard in MT evaluation, these metrics do not fully capture adequacy, faithfulness to visual context, or user-perceived quality. We could also use other automated human-based metrics to complement the current results and provide a more complete assessment.

Finally, some comparisons are constrained by model availability and reproducibility factors. For example, Translatotron-V results were only partially available, and closed-source API-based models may change over time due to backend updates. As a result, exact replication of all reported scores may be difficult.

\section{Conclusion and Future Work}\label{sec:conclusion}

Experimental results demonstrate that \ac{llm}s and \ac{mllm}s are highly effective in translating images containing text. Among the approaches assessed, modular pipelines that combine OCR and LLM-based translation achieved superior performance compared to the end-to-end system, Translatotron-V. This suggests that decomposing the process into specialized stages for detection, recognition, and translation leads to more accurate and reliable outputs. 

Meanwhile, \ac{mllm}s showed strong potential by jointly processing visual and textual information, achieving performance that rivals or exceeds traditional OCR-based pipelines. Their ability to integrate semantic understanding and contextual reasoning enables more flexible and coherent translations, particularly in visually complex scenarios.

Future work will focus on expanding the evaluation to a broader range of models and languages, including low-resource pairs, and exploring hybrid architectures that integrate the strengths of modular and end-to-end approaches. Additionally, the development of larger and more diverse multilingual image datasets would allow for more comprehensive analysis of multi-modal translation performance across domains.

\section*{Acknowledgments}
We would like to express our sincere gratitude to the authors of Translatotron-V \cite{lan2024translatotron} for generously sharing their
models with us.

\label{sec:reference}
\bibliographystyle{apalike}
\bibliography{bibliography}

@article{lan2023exploring,
  title     = {Exploring better text image translation with multimodal codebook},
  author    = {Lan, Zhibin and Yu, Jiawei and Li, Xiang and Zhang, Wen and Luan, Jian and Wang, Bin and Huang, Degen and Su, Jinsong},
  journal   = {arXiv preprint arXiv:2305.17415},
  year      = {2023}
}

@inproceedings{ma2022improving,
  title     = {Improving end-to-end text image translation from the auxiliary text translation task},
  author    = {Ma, Cong and Zhang, Yaping and Tu, Mei and Han, Xu and Wu, Linghui and Zhao, Yang and Zhou, Yu},
  booktitle = {Proceedings of the International Conference on Pattern Recognition},
  pages     = {1664--1670},
  year      = {2022}
}

@inproceedings{qian2024anytrans,
    title = {{A}ny{T}rans: Translate {A}ny{T}ext in the Image with Large Scale Models},
    author = {Qian, Zhipeng  and Zhang, Pei  and Yang, Baosong  and Fan, Kai  and Ma, Yiwei  and
      Wong, Derek F.  and Sun, Xiaoshuai  and Ji, Rongrong},
    booktitle = {Findings of the Association for Computational Linguistics: EMNLP 2024},
    year = {2024},
    pages = {2432--2444},
}

@inproceedings{lan2024translatotron,
  title     = {Translatotron-{V}(ison): An End-to-End Model for In-Image Machine Translation},
  author    = {Lan, Zhibin  and
      Niu, Liqiang  and
      Meng, Fandong  and
      Zhou, Jie  and
      Zhang, Min  and
      Su, Jinsong},
  booktitle = {Findings of the Association for Computational Linguistics: ACL 2024},
  year      = {2024},
  pages     = {5472--5485}
}

@inproceedings{zhu2023peit,
    title = {{PEIT}: Bridging the Modality Gap with Pre-trained Models for End-to-End Image Translation},
    author = {Zhu, Shaolin  and Li, Shangjie  and Lei, Yikun  and Xiong, Deyi},
    booktitle = {Proceedings of the 61st Annual Meeting of the Association for Computational Linguistics},
    year = {2023},
    pages = {13433--13447}
}

@inproceedings{elliott2017imagination,
  title     = "Imagination Improves Multimodal Translation",
  author    = "Elliott, Desmond  and K{\'a}d{\'a}r, {\'A}kos",
  booktitle = "Proceedings of the Eighth International Joint Conference on Natural Language Processing",
  year      = "2017",
  pages     = "130--141",
}

@inproceedings{gao2025multimodal,
  title     = {Multimodal Machine Translation with Text-Image In-depth Questioning},
  author    = {Gao, Yue and Zhao, Jing and Sun, Shiliang and Qiao, Xiaosong and Song, Tengfei and Yang, Hao},
  booktitle = {Findings of the Association for Computational Linguistics: ACL 2025},
  pages     = {9274--9287},
  year      = {2025}
}

@article{lan2025towards,
  title     = {Towards better text image machine translation with multimodal codebook and multi-stage training},
  author    = {Lan, Zhibin and Yu, Jiawei and Liu, Shiyu and Yao, Junfeng and Huang, Degen and Su, Jinsong},
  journal   = {Neural Networks},
  pages     = {107599},
  year      = {2025}
}

@article{ma2023modal,
  title     = {Modal contrastive learning based end-to-end text image machine translation},
  author    = {Ma, Cong and Han, Xu and Wu, Linghui and Zhang, Yaping and Zhao, Yang and Zhou, Yu and Zong, Chengqing},
  journal   = {IEEE/ACM Transactions on Audio, Speech, and Language Processing},
  volume    = {32},
  pages     = {2153--2165},
  year      = {2023}
}

@article{tian2025exploring,
  title     = {Exploring In-Image Machine Translation with Real-World Background},
  author    = {Tian, Yanzhi and Liu, Zeming and Liu, Zhengyang and Guo, Yuhang},
  journal   = {arXiv preprint arXiv:2505.15282},
  year      = {2025}
}

@article{tian2025prim,
  title     = {PRIM: Towards Practical In-Image Multilingual Machine Translation},
  author    = {Tian, Yanzhi and Liu, Zeming and Liu, Zhengyang and Feng, Chong and Li, Xin and Huang, Heyan and Guo, Yuhang},
  journal   = {arXiv preprint arXiv:2509.05146},
  year      = {2025}
}

@inproceedings{tian2023image,
  title     = {In-image neural machine translation with segmented pixel sequence-to-sequence model},
  author    = {Tian, Yanzhi and Li, Xiang and Liu, Zeming and Guo, Yuhang and Wang, Bin},
  booktitle = {Findings of the Association for Computational Linguistics: EMNLP 2023},
  pages     = {15046--15057},
  year      = {2023}
}

@inproceedings{liang2024document,
  title     = {Document image machine translation with dynamic multi-pre-trained models assembling},
  author    = {Liang, Yupu and Zhang, Yaping and Ma, Cong and Zhang, Zhiyang and Zhao, Yang and Xiang, Lu and Zong, Chengqing and Zhou, Yu},
  booktitle = {Proceedings of the 2024 Conference of the North American Chapter of the Association for Computational Linguistics: Human Language Technologies (Volume 1: Long Papers)},
  pages     = {7084--7095},
  year      = {2024}
}

@inproceedings{li2023trocr,
  title     = {Trocr: Transformer-based optical character recognition with pre-trained models},
  author    = {Li, Minghao and Lv, Tengchao and Chen, Jingye and Cui, Lei and Lu, Yijuan and Florencio, Dinei and Zhang, Cha and Li, Zhoujun and Wei, Furu},
  booktitle = {Proceedings of the AAAI conference on artificial intelligence},
  volume    = {37},
  pages     = {13094--13102},
  year      = {2023}
}

@inproceedings{cheng2017focusing,
  title     = {Focusing attention: Towards accurate text recognition in natural images},
  author    = {Cheng, Zhanzhan and Bai, Fan and Xu, Yunlu and Zheng, Gang and Pu, Shiliang and Zhou, Shuigeng},
  booktitle = {Proceedings of the IEEE international conference on computer vision},
  pages     = {5076--5084},
  year      = {2017}
}

@inproceedings{gupta2016synthetic,
  title     = {Synthetic data for text localisation in natural images},
  author    = {Gupta, Ankush and Vedaldi, Andrea and Zisserman, Andrew},
  booktitle = {Proceedings of the IEEE conference on computer vision and pattern recognition},
  pages     = {2315--2324},
  year      = {2016}
}

@inproceedings{karatzas2015icdar,
  title     = {ICDAR 2015 competition on robust reading},
  author    = {Karatzas, Dimosthenis and Gomez-Bigorda, Lluis and Nicolaou, Anguelos and Ghosh, Suman and Bagdanov, Andrew and Iwamura, Masakazu and Matas, Jiri and Neumann, Lukas and Chandrasekhar, Vijay Ramaseshan and Lu, Shijian and others},
  booktitle = {Proceedings of the international conference on document analysis and recognition},
  pages     = {1156--1160},
  year      = {2015},
}

@article{veit2016coco,
  title     = {Coco-text: Dataset and benchmark for text detection and recognition in natural images},
  author    = {Veit, Andreas and Matera, Tomas and Neumann, Lukas and Matas, Jiri and Belongie, Serge},
  journal   = {arXiv preprint arXiv:1601.07140},
  year      = {2016}
}

@inproceedings{wang2011end,
  title     = {End-to-end scene text recognition},
  author    = {Wang, Kai and Babenko, Boris and Belongie, Serge},
  booktitle = {2011 International conference on computer vision},
  pages     = {1457--1464},
  year      = {2011}
}

@article{hameed2025survey,
title       = {Deep Learning Techniques for Machine Translation: A Survey},
journal     = {Procedia Computer Science},
volume      = {258},
pages       = {1022-1037},
year        = {2025},
author      = {Diadeen Ali Hameed and Belal Al-Khateeb},
}

@inproceedings{chatterjee2019findings,
  title     = {Findings of the WMT 2019 Shared Task on Automatic Post-Editing.},
  author    = {Chatterjee, Rajen and Federmann, Christian and Negri, Matteo and Turchi, Marco},
  booktitle = {Proceedings of the Fourth Conference on Machine Translation},
  pages     = {11--28},
  year      = {2019}
}

@inproceedings{farajian2020findings,
  title     = {Findings of the WMT 2020 shared task on chat translation},
  author    = {Farajian, M Amin and Lopes, Ant{\'o}nio V and Martins, Andr{\'e} FT and Maruf, Sameen and Haffari, Gholamreza},
  booktitle = {Proceedings of the Fifth Conference on Machine Translation},
  pages     = {65--75},
  year      = {2020}
}

@inproceedings{specia2021findings,
  title     = {Findings of the WMT 2021 shared task on quality estimation},
  author    = {Specia, Lucia and Blain, Fr{\'e}d{\'e}ric and Fomicheva, Marina and Zerva, Chrysoula and Li, Zhenhao and Chaudhary, Vishrav and Martins, Andr{\'e} FT},
  booktitle = {Proceedings of the Sixth Conference on Machine Translation},
  pages     = {684--725},
  year      = {2021}
}

@inproceedings{wang2024findings,
  title     = {Findings of the {WMT} 2024 Shared Task on Discourse-Level Literary Translation},
  author    = {Wang, Longyue  and Liu, Siyou  and Lyu, Chenyang  and Jiao, Wenxiang  and Wang, Xing  and Xu, Jiahao  and Tu, Zhaopeng  and Gu, Yan  and Chen, Weiyu  and Wu, Minghao  and Zhou, Liting  and Koehn, Philipp  and Way, Andy  and Yuan, Yulin},
  booktitle = {Proceedings of the Ninth Conference on Machine Translation},
  year      = {2024},
  pages     = {699--700},
}

@inproceedings{koehn2005europarl,
  title     = "{E}uroparl: A Parallel Corpus for Statistical Machine Translation",
  author    = "Koehn, Philipp",
  booktitle = "Proceedings of Machine Translation Summit X",
  year      = "2005",
  pages     = "79--86",
}

@inproceedings{tiedemann2020opus,
  title     = {OPUS-MT--Building open translation services for the World},
  author    = {Tiedemann, J{\"o}rg and Thottingal, Santhosh},
  booktitle = {Proceedings of the Annual Conference of the European Association for Machine Translation},
  pages     = {479--480},
  year      = {2020}
}

@article{wang2025rethinking,
  title     = {Rethinking Multilingual Vision-Language Translation: Dataset, Evaluation, and Adaptation},
  author    = {Wang, Xintong and Pan, Jingheng and Liu, Yixiao and Zhao, Xiaohu and Lyu, Chenyang and Wu, Minghao and Biemann, Chris and Wang, Longyue and Xu, Linlong and Luo, Weihua and others},
  journal   = {arXiv preprint arXiv:2506.11820},
  year      = {2025}
}

@inproceedings{li2025mit10m,
    title = {{MIT}-10{M}: A Large Scale Parallel Corpus of Multilingual Image Translation},
    author = {Li, Bo  and Zhu, Shaolin  and Wen, Lijie},
    booktitle = {Proceedings of the 31st International Conference on Computational Linguistics},
    year = {2025},
    pages = {5154--5167},
}

@article{shi2016end,
  author    = {Shi, Baoguang and Bai, Xiang and Yao, Cong},
  journal   = {IEEE Transactions on Pattern Analysis and Machine Intelligence}, 
  title     = {An End-to-End Trainable Neural Network for Image-Based Sequence Recognition and Its Application to Scene Text Recognition}, 
  year      = {2017},
  volume    = {39},
  number    = {11},
  pages     = {2298--2304}
}

@inproceedings{morris2004and,
  title={From WER and RIL to MER and WIL: improved evaluation measures for connected speech recognition.},
  author={Morris, Andrew Cameron and Maier, Viktoria and Green, Phil D},
  booktitle={Proceedings of Interspeech},
  pages={2765--2768},
  year={2004}
}

@inproceedings{post2018call,
    title = {A Call for Clarity in Reporting {BLEU} Scores},
    author = {Post, Matt},
    booktitle = {Proceedings of the Third Conference on Machine Translation: Research Papers},
    year = {2018},
    pages = {186--191},
}

@article{fan2021beyond,
  title={Beyond english-centric multilingual machine translation},
  author={Fan, Angela and Bhosale, Shruti and Schwenk, Holger and Ma, Zhiyi and El-Kishky, Ahmed and Goyal, Siddharth and Baines, Mandeep and Celebi, Onur and Wenzek, Guillaume and Chaudhary, Vishrav and others},
  journal={Journal of Machine Learning Research},
  volume={22},
  number={107},
  pages={1--48},
  year={2021}
}

@article{dubey2024llama,
  title={The llama 3 herd of models},
  author={Dubey, Abhimanyu and Jauhri, Abhinav and Pandey, Abhinav and Kadian, Abhishek and Al-Dahle, Ahmad and Letman, Aiesha and Mathur, Akhil and Schelten, Alan and Yang, Amy and Fan, Angela and others},
  journal={arXiv e-prints},
  year={2024}
}

@article{martins2025eurollm,
  title={Eurollm-9b: Technical report},
  author={Martins, Pedro Henrique and Alves, Jo{\~a}o and Fernandes, Patrick and Guerreiro, Nuno M and Rei, Ricardo and Farajian, Amin and Klimaszewski, Mateusz and Alves, Duarte M and Pombal, Jos{\'e} and Boizard, Nicolas and others},
  journal={arXiv preprint arXiv:2506.04079},
  year={2025}
}

@article{comanici2025gemini,
  title={Gemini 2.5: Pushing the frontier with advanced reasoning, multimodality, long context, and next generation agentic capabilities},
  author={Comanici, Gheorghe and Bieber, Eric and Schaekermann, Mike and Pasupat, Ice and Sachdeva, Noveen and Dhillon, Inderjit and Blistein, Marcel and Ram, Ori and Zhang, Dan and Rosen, Evan and others},
  journal={arXiv preprint arXiv:2507.06261},
  year={2025}
}

@article{Hochreiter97,
  title={Long short-term memory},
  author={Hochreiter, Sepp and Schmidhuber, J{\"u}rgen},
  journal={Neural computation},
  volume={9},
  number={8},
  pages={1735--1780},
  year={1997},
  publisher={MIT press}
}

@article{liu2023pre,
  title={Pre-train, prompt, and predict: A systematic survey of prompting methods in natural language processing},
  author={Liu, Pengfei and Yuan, Weizhe and Fu, Jinlan and Jiang, Zhengbao and Hayashi, Hiroaki and Neubig, Graham},
  journal={ACM computing surveys},
  volume={55},
  number={9},
  pages={1--35},
  year={2023}
}

@article{simonyan2014very,
  title={Very deep convolutional networks for large-scale image recognition},
  author={Simonyan, Karen and Zisserman, Andrew},
  journal={arXiv preprint arXiv:1409.1556},
  year={2014}
}

@inproceedings{wolf2020transformers,
  title={Transformers: State-of-the-art natural language processing},
  author={Wolf, Thomas and Debut, Lysandre and Sanh, Victor and Chaumond, Julien and Delangue, Clement and Moi, Anthony and Cistac, Pierric and Rault, Tim and Louf, Remi and Funtowicz, Morgan and others},
  booktitle={Proceedings of the 2020 conference on empirical methods in natural language processing: system demonstrations},
  pages={38--45},
  year={2020}
}

@misc{doctr2021,
    title={docTR: Document Text Recognition},
    author={Mindee},
    year={2021},
    publisher = {GitHub},
    howpublished = {\url{https://github.com/mindee/doctr}}
}

@inproceedings{mohammadshahi2022-mall,
    title = "{SM}a{LL}-100: Introducing Shallow Multilingual Machine Translation Model for Low-Resource Languages",
    author = "Mohammadshahi, Alireza  and
      Nikoulina, Vassilina  and
      Berard, Alexandre  and
      Brun, Caroline  and
      Henderson, James  and
      Besacier, Laurent",
    booktitle = "Proceedings of the 2022 Conference on Empirical Methods in Natural Language Processing",
    year = "2022",
    pages = "8348--8359",
}

@inproceedings{boyd2025machine,
    title = "Machine Translation in the {AI} Era: Comparing previous methods of machine translation with large language models",
    author = "Boyd, William Jock  and
      Mitkov, Ruslan",
    booktitle = "Proceedings of the 8th Workshop on Challenges and Applications of Automated Extraction of Socio-political Events from Texts",
    year = "2025",
    pages = "38--51",
}

\end{document}